\documentclass[10pt,twocolumn,letterpaper]{article}

\usepackage[pagenumbers]{cvpr}              

\usepackage{graphicx}
\usepackage{amsmath}
\usepackage{amssymb}
\usepackage{booktabs}
\usepackage{multirow}
\usepackage{xcolor,colortbl}   
\usepackage{pifont}   
\usepackage[ruled,vlined]{algorithm2e}
\usepackage{rotating}
\usepackage{float}
\usepackage[accsupp]{axessibility}

\definecolor{Green}{RGB}{0, 180, 0}
\definecolor{Red}{RGB}{180, 0, 0}
\definecolor{Blue}{RGB}{30, 0, 180}
\definecolor{Gray}{gray}{0.9}
\definecolor{Textgray}{gray}{0.4}
\newcommand{\cmark}{{\textcolor{Green}{\ding{51}}}}%
\newcommand{\xmark}{{\textcolor{Red}{\ding{55}}}}%

\definecolor{cvprblue}{rgb}{0.21,0.49,0.74}
\usepackage[pagebackref,breaklinks,colorlinks,allcolors=cvprblue]{hyperref}

\def\paperID{35415} 
\def\confName{CVPR}
\def\confYear{2026}

\title{Neural Distribution Prior for LiDAR Out-of-Distribution Detection}

\author{Zizhao Li
\and
Zhengkang Xiang
\and
Jiayang Ao
\and
Feng Liu
\and
Joseph West
\quad
Kourosh Khoshelham
\\
The University of Melbourne, Parkville, Victoria 3010, Australia
}

\begin{document}
\maketitle
\begin{abstract}
LiDAR-based perception is critical for autonomous driving due to its robustness to poor lighting and visibility conditions. Yet, current models operate under the closed-set assumption and often fail to recognize unexpected out-of-distribution (OOD) objects in the open world. Existing OOD scoring functions exhibit limited performance because they ignore the pronounced class imbalance inherent in LiDAR OOD detection and assume a uniform class distribution. To address this limitation, we propose the Neural Distribution Prior (NDP), a framework that models the distributional structure of network predictions and adaptively reweights OOD scores based on alignment with a learned distribution prior. NDP dynamically captures the logit distribution patterns of training data and corrects class-dependent confidence bias through an attention-based module. We further introduce a Perlin noise–based OOD synthesis strategy that generates diverse auxiliary OOD samples from input scans, enabling robust OOD training without external datasets. Extensive experiments on the SemanticKITTI and STU benchmarks demonstrate that NDP substantially improves OOD detection performance, achieving a point-level AP of 61.31\% on the STU test set, which is more than 10$\times$ higher than the previous best result. Our framework is compatible with various existing OOD scoring formulations, providing an effective solution for open-world LiDAR perception. \footnote{Project Page: https://cs-lzz.github.io/ndp-demo}
\end{abstract}    
\section{Introduction}
\label{sec:intro}

LiDAR sensing plays an important role in autonomous driving, providing precise 3D information about the surrounding environment.
Its robustness to illumination and weather variations, together with its geometric precision, make LiDAR a fundamental sensing modality for scene understanding and safe navigation~\cite{Bijelic_2020_STF,li2025outofdistributiondetection3dapplications}.
However, autonomous vehicles must operate in open-world environments where unknown or unexpected objects may appear at any time.
Such out-of-distribution (OOD) objects, including fallen branches, construction machinery, or road debris, occur rarely but can have severe safety consequences.

\begin{figure}[t]
    \centering
    \includegraphics[width=0.95\linewidth]{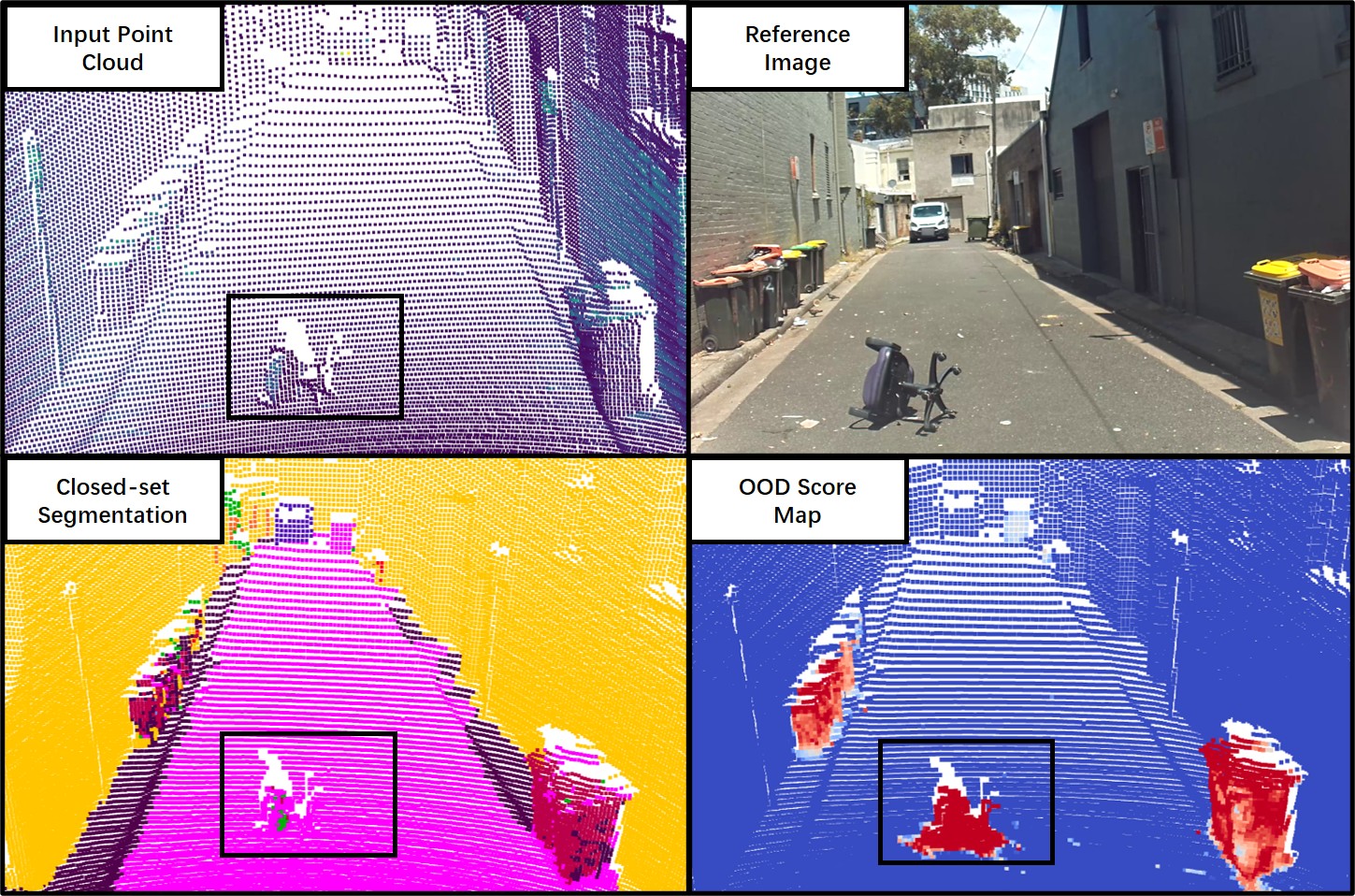}
    \caption{OOD objects are hazardous for LiDAR perception models because they are often misclassified as known categories.  
For example, an armchair on the road is incorrectly predicted as “road” by a closed-set model. The proposed NDP correctly assigns high OOD scores (red) to the road hazard (armchair) and to the roadside anomalies (rubbish bins).
}
    \label{fig:ood_chair}
    \vspace{-0.5cm}
\end{figure}

However, scene perception models often operate under the closed-set assumption, which forces them to assign one of the known labels to OOD objects, leading to incorrect and potentially unsafe predictions~\cite{fishyscapes_2,yang2024generalizedood,tosr14}.
As shown in \cref{fig:ood_chair}, a closed-set LiDAR segmentation model misclassifies an armchair on the road as part of the road surface, while a reliable LiDAR perception model should distinguish such OOD objects from known classes, as illustrated in the bottom-right figure.
Reliable OOD detection is therefore crucial for ensuring safe and dependable LiDAR perception in real-world driving scenarios.

Although 2D OOD detection has been extensively studied~\cite{grcic2022densehybrid,rpl2023,tian2022pebal,chan2021entropy,nayal2023rba,Li_2025_BMVC}, extending these approaches to LiDAR remains challenging because point clouds are inherently sparse, irregular, and prone to occlusion~\cite{kosel2024mmod3d}. LiDAR datasets also exhibit severe class imbalance: dominant categories such as road and building contain most points, while traffic participants such as bicycles are sparsely represented, making OOD detection even more challenging.

Class imbalance is a fundamental issue in real-world perception~\cite{cao2019imbalanced,park2022imbalanced}.
Models trained on skewed data tend to favor dominant classes, leading to higher inter-class confusion risks~\cite{jiang2023classprior,choi2023balanced}.
Many existing OOD detection methods~\cite{hendrycks2018baseline,huang2021importance,energy,odin} assume roughly uniform class distributions and rely on fixed scoring functions.
In large-scale LiDAR scenes, per-class point counts can vary by several orders of magnitude~\cite{behley2019semantickitti,caesar2020nuscenes}, causing static OOD scores to overfit frequent classes and fail on tail classes~\cite{jiang2023classprior,liu2024imood,choi2023balanced}.
Furthermore, the number of auxiliary OOD samples introduced during training is also small compared to the vast number of in-distribution points.
We observe that dataset-level class priors~\cite{jiang2023classprior,liu2024imood} are insufficient to correct the bias introduced by severe class imbalance in LiDAR data.

Therefore, we propose the Neural Distribution Prior (NDP), a learnable module that models inter-class relationships to more accurately characterize the network’s predictive distribution.
NDP projects the output logits of each sample into a latent embedding space and performs cross-attention with a learnable distribution prior matrix to capture distributional relationships across classes. This module models the typical behavior of the network’s predictions during training and serves as a reference distribution that regularizes model outputs, improving calibration and robustness under class-imbalanced LiDAR scenes.

Another challenge is the model’s unawareness of OOD objects.
Outlier Exposure (OE)~\cite{hendrycks2019oe} addresses this issue by introducing auxiliary OOD samples during training, and has proven effective for both image-based~\cite{rpl2023,grcic2022densehybrid,tian2022pebal,chan2021entropy,wang2023outofdistribution,zheng_atol,wangDAOL} and LiDAR-based tasks~\cite{cen2021real,li2025relativeenergylearninglidar,lion2025}.
By introducing auxiliary OOD data during training, OE encourages the model to assign low confidence to unfamiliar inputs and to learn more discriminative decision boundaries, where OOD samples typically come from external datasets.  
However, extending this idea to LiDAR perception is non-trivial.  
Point clouds are sparse, irregular, and strongly affected by occlusion~\cite{kosel2024mmod3d}, making the use of external datasets challenging and requiring labor-intensive adaptation to maintain geometric and domain consistency.

An alternative is to exploit points within the existing datasets that are excluded from the closed-set training classes, a strategy known as void classification~\cite{fishyscapes_2,nekrasov2025stu}.
While appealing in principle, our observations show that such points are not a reliable OOD source.
Their diversity is limited, and many correspond to regions with meaningful but unlabeled semantics rather than true anomalies.
Models trained on these points easily overfit, leading to poor generalization to diverse OOD objects.

To generate diverse and realistic auxiliary OOD samples,  we propose a Perlin noise–based OOD synthesis strategy that generates diverse pseudo-OOD samples directly from inlier point clouds.
By perturbing local surface geometry with smooth fractal noise fields, the method introduces realistic variations in shape and contour while preserving the global semantic layout. 
In addition, it alleviates the need for external datasets and complex post-processing.

To exploit the void class as an auxiliary OOD source, we propose Soft Outlier Exposure (SOE).
Instead of treating void points as fully reliable OOD samples, SOE assigns them soft OOD labels that reflect their uncertain nature.
This treatment allows the model to learn from these ambiguous regions while preventing overfitting to certain object categories.

We evaluate NDP on the SemanticKITTI~\cite{behley2019semantickitti} and STU~\cite{nekrasov2025stu} benchmarks.
Our method achieves state-of-the-art OOD detection performance while maintaining strong in-distribution accuracy.

Our key contributions are summarized as follows:
\begin{itemize}
    \item We introduce the Neural Distribution Prior (NDP), a learnable prior estimation module that models the distributional structure of network predictions and adaptively adjusts OOD scores to improve calibration under class imbalance.  
Combined with the Extended Energy score proposed in this work, our method achieves 61.31\% AP on the STU test set, which is over $10\times$ higher than the previous SOTA result.
    \item We develop a Perlin noise–based OOD synthesis method that generates diverse synthetic OOD samples directly from inlier scans, providing additional negative supervision without external datasets.  
    \item We propose a Soft Outlier Exposure (SOE) training strategy that jointly leverages synthetic OOD samples and unreliable void regions by assigning soft OOD labels, enabling stable optimization and better generalization.
\end{itemize}

\section{Related Work}
\paragraph{OOD Detection}
OOD detection aims to identify test inputs that deviate from the training distribution, allowing models to abstain from overconfident predictions on unknown data~\cite{yang2024generalizedood}.  
Early OOD detection research primarily targeted image classification tasks~\cite{hendrycks2018baseline,hendrycks2019oe,odin,energy,md,watermark}.  
Subsequent works extended this problem to dense prediction settings, giving rise to pixel-wise OOD detection or anomaly segmentation.  
Benchmarks such as Fishyscapes~\cite{blum2021fishyscapes} and SMIYC~\cite{chan2021segmentmeifyoucan} have facilitated this transition.  
Most early anomaly segmentation methods~\cite{hendrycks2018baseline,chan2021entropy,lakshminarayanan2017deepensemble,tian2022pebal,grcic2022densehybrid,liang2022gmmseg,rpl2023} are built upon convolutional semantic segmentation backbones such as DeepLab~\cite{chen2018deeplabv3}, where OOD scores are derived from pixel-level softmax probabilities, entropy, or energy functions.
More recent approaches have shifted toward transformer-based segmentation architectures, particularly those inspired by MaskFormer~\cite{cheng2021maskformer} and Mask2Former~\cite{cheng2021mask2former}.
Several subsequent studies~\cite{delic2024uno,nayal2023rba,rai2023mask2anomaly} extend these frameworks to anomaly segmentation by refining the model architecture and incorporating object-level reasoning.


\begin{figure*}[ht]
    \centering
    \includegraphics[width=\linewidth]{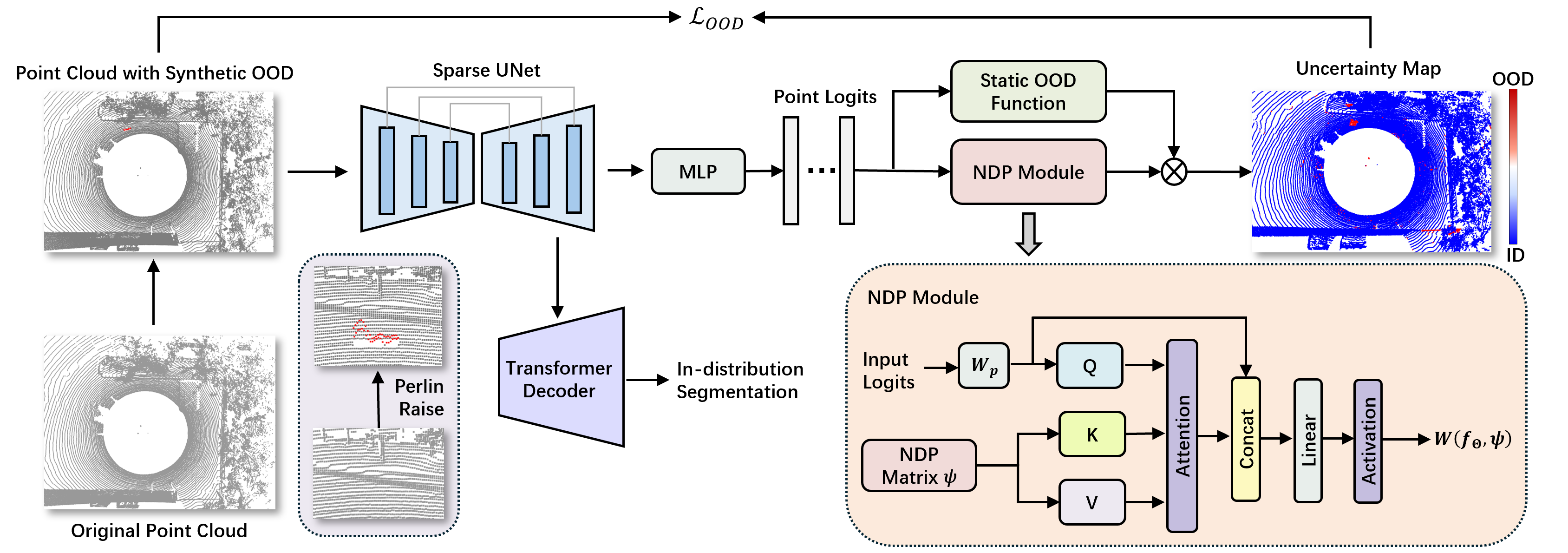}
    \caption{Overview of the proposed Neural Distribution Prior (NDP) framework.  
Given an input point cloud, synthetic OOD samples are generated using the Perlin Raise procedure and jointly trained with in-distribution data.  
A sparse UNet~\cite{choy2019minkowski} extracts point features, which are processed by an MLP to produce class logits for OOD detection, while the transformer decoder utilizes multi-level features from the UNet to generate mask-based predictions for in-distribution segmentation.  
The NDP module projects the input logits into a latent embedding space and performs cross-attention with a learnable prior matrix $\psi$, producing a reweighting term $W(f_{\Theta}, \psi)$ that adjusts the static OOD score to generate calibrated uncertainty maps.}
    \label{fig:framework}
    \vspace{-0.3cm}
\end{figure*}

\paragraph{LiDAR OOD Detection}
In contrast, OOD detection in 3D LiDAR data remains far less explored.  
Early studies~\cite{mood3D,kosel2024mmod3d} adapted OOD scoring methods to pretrained 3D object detectors and evaluated their post-hoc performance on benchmarks such as KITTI~\cite{geiger2013KITTI} and nuScenes~\cite{caesar2020nuscenes}.  
For LiDAR semantic segmentation, Cen \etal~\cite{cen2021real} proposed REAL, which introduces auxiliary logits to learn pseudo-OOD representations generated by scaling point clouds. Li \etal~\cite{apf2023} employed adversarial prototypes to perform feature-level OOD learning. Xu \etal~\cite{lion2025} developed LiON, which synthesizes diverse outlier shapes from ShapeNet~\cite{shapenet2015} and learns a point-wise abstaining margin with a selective classification framework.

Many existing approaches rely on reusing autonomous driving datasets~\cite{caesar2020nuscenes,behley2019semantickitti}, where minority or void classes are treated as OOD samples.  
However, this design does not reflect realistic road hazards and limits the diversity of OOD instances.  
To address this limitation, Nekrasov \etal~\cite{nekrasov2025stu} introduced the STU benchmark, which provides 3D LiDAR data with annotations of real road hazards, and established Mask4Former~\cite{yilmaz2024mask4former} with post-hoc OOD scores as the baseline.  
Subsequent work~\cite{li2025relativeenergylearninglidar} employed energy-based scoring and OOD synthesis to improve detection performance.

\paragraph{OOD Detection under Imbalanced Data Distribution}
Class imbalance is inherent in real-world data, yet studies on OOD detection under imbalanced conditions remain limited.
Liu~\etal~\cite{openlongtailrecognition} investigate open-set recognition under class imbalance, while Wang~\etal~\cite{wang2022partial} propose an asymmetric contrastive learning framework that mitigates performance degradation on imbalanced data.  
Subsequent studies~\cite{Wei2024,Miao2024,sapkota2023adaptive,choi2023balanced} explore abstention learning, distributionally robust optimization, and adaptive regularization strategies to handle imbalance in OOD detection.  
Jiang~\etal~\cite{jiang2023classprior} propose class-prior reweighting for post-hoc normalization, but their method is constrained to pretrained models.  
Liu~\etal~\cite{liu2024imood} proposed a statistical framework that unifies training-time regularization to mitigate class bias and improve OOD detection under imbalanced data distributions.
More recent work~\cite{darl2025} address the gradient conflicts in long-tailed OOD detection.

\section{Method}

\subsection{Problem Definition}

OOD detection aims to learn a classifier $G$ such that, for any test input $x$ drawn from either the in-distribution $P_{X^{in}}$ or the OOD $P_{X^{out}}$,  
if $x \sim P_{X^{in}}$, the classifier $G$ correctly assigns $x$ to one of the inlier classes in $\mathcal{Y}^{in}$;  
and if $x \sim P_{X^{out}}$, $G$ identifies $x$ as OOD data~\cite{fang2022ooddetection}.

In LiDAR semantic segmentation, each input $x$ is a point cloud $X = \{x_j\}_{j=1}^{M}$,  
and the detector computes a point-wise score $S(f_{\Theta},x_j)$ to distinguish  
in-distribution (ID) and OOD points:
\begin{equation}
G(x_j)=
\begin{cases}
\text{ID}, & S(f_{\Theta},x_j)\le\gamma,\\[2pt]
\text{OOD}, & S(f_{\Theta},x_j) > \gamma,
\end{cases}
\end{equation}
where $\gamma$ is a decision threshold.

\subsection{General Architecture}
As shown in~\cref{fig:framework}, our model is built upon the Mask4Former-3D framework~\cite{yilmaz2024mask4former}, which integrates a transformer decoder with a multi-scale sparse UNet~\cite{choy2019minkowski} encoder for 3D panoptic segmentation.
We use the point features extracted from the sparse UNet to predict class logits, from which the OOD score and neural weighting function are computed to produce the final point-wise OOD score.  
In parallel, the transformer decoder is trained in the standard closed-set setting to preserve the panoptic segmentation capability.
This design enhances OOD detection performance and simultaneously maintains strong in-distribution segmentation capability.

\subsection{Perlin Noise-based OOD Synthesis}

Introducing auxiliary data is a common practice in LiDAR OOD detection~\cite{cen2021real,nekrasov2025stu,lion2025}, but it presents several challenges.  
Using external datasets increases complexity and requires careful alignment with the target domain, while the auxiliary samples must be sufficiently diverse to prevent overfitting and ensure robust generalization.

To overcome these challenges and generate diverse and generalizable auxiliary samples, we propose a simple method that involves perturbing the surface geometry of in-distribution point clouds using Perlin Noise.
Perlin Noise~\cite{Perlin1985287} is a smooth, spatially coherent noise function widely used in graphics and simulation for generating natural textures,  
and has also proven effective for synthesizing structural defects in industrial anomaly detection~\cite{ZAVRTANIK2024113,li2025das3ddualmodalityanomalysynthesis,cheng20253dpnas3dindustrialsurface,Tao_2025_ICCV}.  
The proposed \emph{Perlin Raise} algorithm (Algorithm~\ref{alg:perlin_raise}) 
generates spatially coherent surface perturbations that simulate realistic road anomalies.  
Given a LiDAR frame $(P, L)$, we sample a road patch of radius $r$, 
generate a Perlin field $n(u,v)$ over the patch, 
and assign each point a noise value $n_i$.  
Points within the top $\rho$ fraction are selected, 
their noise values are locally normalized to obtain gain $g_i \in [0,1]$, 
and each point is elevated by $\Delta z_i = \alpha \cdot g_i$.
Points are then clustered via DBSCAN~\cite{dbscan}, 
and the largest connected component is labeled as OOD.

\begin{algorithm}[t]
\caption{Perlin Raise Algorithm}
\label{alg:perlin_raise}
\SetAlgoLined
\DontPrintSemicolon
\KwIn{Point cloud $P$, labels $L$, 
patch radius $r$, noise strength $\alpha$, target ratio $\rho$}
\KwOut{Modified $P'$, $L'$}
\tcp{Sample Random Flat Region}
$c \leftarrow$ random road point\;
$\mathcal{N} \leftarrow$ KDTree($P$).query\_ball\_point($c$, $r$)\;
\tcp{Compute Perlin Grid}
$n \leftarrow$ PerlinField$(X=\{P_i \mid i \in \mathcal{N}\})$\;
\tcp{Sample per-point Perlin}
$n_i \leftarrow$ SamplePerlin$(P_i)$ for $i \in \mathcal{N}$\;
\tcp{Select Raised Region}
$\mathcal{B} \leftarrow \{i \in \mathcal{N} \mid n_i > \text{quantile}(n, 1-\rho)\}$\; 
$g_i \leftarrow Normalize(n_i)$ for $i \in \mathcal{B}$\;
$\Delta z_i \leftarrow \alpha \cdot g_i$ for $i \in \mathcal{B}$\;
\tcp{Filter the largest cluster}
$\mathcal{C} \leftarrow \textsc{DBSCAN}(P[\mathcal{B}])$\;
$k \leftarrow \textsc{Mode}(\mathcal{C})$ \tcp{largest cluster index}
$P'_i.z \leftarrow P_i.z + \Delta z_i$ for $i \in \mathcal{C}_k$\;
$L'_i \leftarrow \text{OOD}$ for $i \in \mathcal{C}$\;
\Return $P', L'$\;
\end{algorithm}

\subsection{Neural Distribution Prior}


While Perlin Raise expands the diversity of OOD supervision, handling class imbalance and confidence bias remains challenging. 
To tackle this, we propose a data-driven and learnable prior estimation framework called the \emph{Neural Distribution Prior (NDP)}.  
Instead of relying on a fixed OOD scoring function, NDP models the distributional relationships among predictions and adaptively calibrates the output OOD score.
For an arbitrary baseline OOD score $S_{\text{Method}}(f_{\Theta},x)$,  
we define the NDP reweighted score as
\begin{equation}
S_{\text{NDP}}(f_{\Theta},x)
= S_{\text{Method}}(f_{\Theta},x)
\cdot W(f_{\Theta}, \psi),
\label{eq:w_ndp}
\end{equation}
where $W(f_{\Theta}, \psi)$ denotes a neural weighting function that reflects how well each prediction aligns with the learned distribution prior $\psi$.  

Given network logits $f_{\Theta}(x) \in \mathbb{R}^{N \times K}$, where $N$ is the number of points and $K$ is the length of a logit vector, the neural reweighting function $W(f_{\Theta}, \psi)$ first projects each logit vector into a latent embedding $e = W_p f_{\Theta}(x)$,
where $W_p$ is a learnable linear projection and $d$ is the latent dimension.  
A learnable prior table $\psi \in \mathbb{R}^{K \times d}$ stores class-level embeddings that capture the characteristic distribution of each class in the training data and is optimized jointly with the network through gradient-based learning.  
To capture relationships between sample embeddings and the learned prior, 
we compute a cross-attention between $e$ and $\psi$ as
\begin{equation}
z = \mathrm{softmax}\!\left(\frac{Q(e)\,K(\psi)^{\top}}{\sqrt{d}}\right)V(\psi),
\end{equation}
where $Q(\cdot)$, $K(\cdot)$, and $V(\cdot)$ are linear projections that generate query, key, and value representations, respectively.  
The resulting context $z \in \mathbb{R}^{N \times d}$ encodes how each prediction aligns with the learned prior manifold.  
A linear mapping then converts the concatenated embedding $[e, z]$ into a scalar weight $w = \mathrm{ReLU}(W_s [e, z]) + 1$.  
The resulting $w$ serves as the neural prior term $W(f_{\Theta}, \psi)$ in~\cref{eq:w_ndp}, 
modulating the OOD energy or confidence of each prediction according to its alignment with $\psi$.

In this way, our method captures the underlying patterns of the logit distribution and dynamically calibrates the OOD score. 
Moreover, it is flexible and can be integrated with various OOD scoring functions, including entropy~\cite{chan2021entropy}, energy~\cite{energy}, and extended energy, to improve robustness under class imbalance.

\paragraph{NDP Reweighted Entropy}
One possible choice of OOD score is entropy. Entropy-based OOD detection measures confidence via the softmax entropy of the network outputs~\cite{chan2021entropy}. 
We define the
\emph{NDP Reweighted Entropy} score as:
\begin{equation}
S_{\text{NDP-Entropy}}(f_{\Theta}, x)
\;=\; - \sum_{i=1}^{K} p_i \log p_i \cdot W(f_{\Theta}, \psi),
\label{eq:ndp_entropy}
\end{equation}
where $p_i$ is the softmax probability of each class.

\paragraph{NDP Reweighted Energy}
Energy-based OOD detection~\cite{energy,du2021vos,tian2022pebal} computes the confidence of a sample
using the negative log-sum-exp of the logits.
The NDP reweighted energy score is defined as
\begin{equation}
S_{\text{NDP-Energy}}(f_{\Theta}, x)
= -\log \sum_{i=1}^{K}
   e^{f_{\Theta}^{i}(x)} \cdot W(f_{\Theta}, \psi).
\label{eq:ndp_energy}
\end{equation}

\paragraph{NDP Reweighted Extended Energy}
A common strategy in LiDAR OOD detection is to allocate additional negative logits for modeling OOD samples~\cite{cen2021real,lion2025}.
Recent energy-based OOD detectors~\cite{dual_energy} similarly incorporate negative logits to represent auxiliary OOD data. 
Introducing negative logits enables fine-grained partitioning of the logit space, reducing false positives~\cite{jiang2024negative} and allowing explicit modeling of OOD samples.

Building on this idea, we propose \emph{Extended Energy}, which incorporates logits for both ID and OOD samples to facilitate more effective learning. 
Given an input point $x$, the network outputs logits $f(x) \in \mathbb{R}^{2K}$, where the first $K$ channels correspond to ID classes ($y^+$) and the remaining $K$ channels serve as their negative OOD counterparts ($y^-$). 
Let $y = y^+ \cup y^-$ denote the complete set of channels. 
We define the \emph{NDP Reweighted Extended Energy (NDP-EE)} score as
\begin{equation}
S_{\text{NDP-EE}}(f_{\Theta}, x)
= \log 
\frac{\sum_{i \in y} \exp(f_{\Theta}^{i}(x))}
     {\sum_{i \in y^+} \exp(f_{\Theta}^{i}(x))} 
\cdot W(f_{\Theta}, \psi).
\label{eq:ndp_re}
\end{equation}

This formulation explicitly models the network’s behavior under OOD inputs, leading to improved separation between ID and OOD samples.


\begin{table*}[t]
    \centering
    \caption{Anomaly segmentation performance on the validation set of the STU benchmark. All methods use the Mask4Former~\cite{yilmaz2024mask4former} architecture. NDP substantially improves both point-level and object-level OOD detection performance compared with existing methods.}
    \vspace{-0.1cm}
    \tiny
    \resizebox{\textwidth}{!}{
    \begin{tabular}{lcccccccccc}
        \toprule
        \multirow{2}{*}{Method} & \multirow{2}{*}{\shortstack{Auxiliary \\ OOD Data}} & \multicolumn{3}{c}{Point-Level OOD} && \multicolumn{5}{c}{Object-Level OOD} \\
        \cline{3-5} \cline{7-11}
         & & AUROC~$\uparrow$ & FPR@95~$\downarrow$ & AP~$\uparrow$ && RecallQ~$\uparrow$ & SQ~$\uparrow$ & RQ~$\uparrow$ & UQ~$\uparrow$ & PQ~$\uparrow$ \\
        \midrule
        Deep Ensemble~\cite{lakshminarayanan2017deepensemble} & \xmark & 90.93 & 37.34 & \cellcolor{gray!20}6.94 && 17.70 & 79.96 & 9.10 & 14.15 & \cellcolor{gray!20}7.27 \\
        MC Dropout~\cite{srivastava2014mcdropout}    & \xmark       & 65.76 & 79.82 & \cellcolor{gray!20}0.17 && 3.54 & 74.36 & 3.48 & 2.63 & \cellcolor{gray!20}2.59 \\
        MaxLogit~\cite{hendrycks2022streethazards}      & \xmark     & 87.27 & 68.76 & \cellcolor{gray!20}2.02 && 26.64 & 79.26 & 2.06 & 21.12 & \cellcolor{gray!20}1.63 \\
        Void Classifier~\cite{blum2021fishyscapes} & \cmark & 89.77 & 79.50 & \cellcolor{gray!20}2.62 && 17.35 & \textbf{81.27} & 8.98 & 14.10 & \cellcolor{gray!20}7.30 \\
        RbA~\cite{nayal2023rba}                  & \xmark    & 73.00 & 100.0 & \cellcolor{gray!20}1.64 && 21.84 & 78.58 & 2.75 & 17.16 & \cellcolor{gray!20}2.16 \\
        \midrule
        NDP-Energy & \cmark & 99.37 & 2.35 & \cellcolor{gray!20}66.54 && 26.80 & 71.25 & 36.08 & 19.10 & \cellcolor{gray!20}25.71 \\
        NDP-Entropy & \cmark & 98.85 & 6.89 & \cellcolor{gray!20}27.90 &&  33.60 & 72.28 & 22.66 & 24.28 & \cellcolor{gray!20}16.38 \\
        NDP-EE & \cmark & \textbf{99.53} & \textbf{1.43} & \cellcolor{gray!20}   \textbf{74.24} && \textbf{50.50} & 74.11 & \textbf{52.13} & \textbf{37.42} & \cellcolor{gray!20}\textbf{38.63} \\
        \bottomrule
    \end{tabular}}
    \label{tab:val-results}
\end{table*}

\begin{table*}[t]
    \centering
    \caption{Anomaly segmentation performance of the test set of the STU benchmark. All methods use the Mask4Former~\cite{yilmaz2024mask4former} architecture. NDP consistently achieves state-of-the-art performance under both point-level and object-level evaluations.
}
\vspace{-0.1cm}
    \tiny
    \resizebox{\textwidth}{!}{
    \begin{tabular}{lccccccccccc}
        \toprule
        \multirow{2}{*}{Method} & \multirow{2}{*}{\shortstack{Auxiliary \\ OOD Data}} & \multicolumn{3}{c}{Point-Level OOD} && \multicolumn{5}{c}{Object-Level OOD} \\
        \cline{3-5}
        \cline{7-11}
         &  & AUROC~$\uparrow$ & FPR@95~$\downarrow$ & AP~$\uparrow$ && RecallQ~$\uparrow$ & SQ~$\uparrow$ & RQ~$\uparrow$ & UQ~$\uparrow$ & PQ~$\uparrow$ \\
        \midrule
        Deep Ensemble~\cite{lakshminarayanan2017deepensemble}         & \xmark  & 86.74 & 58.05 & \cellcolor{gray!20}{5.17} && 16.75 & {84.49} & 10.43 & 14.16 & \cellcolor{gray!20}{8.81} \\
        MC Dropout~\cite{srivastava2014mcdropout}       & \xmark  & 61.51 & 82.37 &  \cellcolor{gray!20}{0.11} && 
        2.25 & \textbf{86.72} & 1.95 & 2.14 & \cellcolor{gray!20}{1.86} \\
        MaxLogit~\cite{hendrycks2022streethazards}        & \xmark  & 84.53 & 81.49 & \cellcolor{gray!20}{0.95} && \textbf{26.14} & 83.06 & 2.13 & 21.71 & \cellcolor{gray!20}{1.77} \\
        Void Classifier~\cite{blum2021fishyscapes}  & \cmark  & {85.99} & {78.60} & \cellcolor{gray!20}{{3.92}} && 17.64 & 84.40 & {8.19} & 14.89 & \cellcolor{gray!20}{{6.91}} \\
        RbA~\cite{nayal2023rba}             & \xmark  & 66.38 & 100.0 & \cellcolor{gray!20}{0.81} && {24.04} & 83.28 & 3.23 & {20.02} & \cellcolor{gray!20}{2.69} \\
        \midrule
        NDP-Entropy & \cmark & 98.41 & 9.83 & \cellcolor{gray!20}29.53 && 16.33 & 74.40 & 15.09 & 12.15 & \cellcolor{gray!20}11.22 \\
        NDP-Energy & \cmark & 99.21 & 3.65 & \cellcolor{gray!20}53.75 && 20.31 & 78.21 & 26.94 & 15.88 & \cellcolor{gray!20}21.07 \\ 
        NDP-EE & \cmark  & \textbf{99.26} & \textbf{3.30} & \cellcolor{gray!20}{\textbf{61.31}} && 25.58 & 79.93 & \textbf{31.26} & \textbf{20.44} & \cellcolor{gray!20}{\textbf{24.99}} \\
        \bottomrule
    \end{tabular}}
    \label{tab:test-results}
    \vspace{-0.2cm}
\end{table*}

\subsection{Training Objective}


OOD training is often formulated either as a regularization term~\cite{energy,choi2023balanced} or as a binary classification problem~\cite{du2021vos,li2025relativeenergylearninglidar}. 
A key insight of our approach is that not all auxiliary OOD samples contribute equally to the training process.  
Void regions, such as background points or outliers that exist in the raw training data but are excluded from the closed-set labels, are often used as auxiliary OOD samples~\cite{fishyscapes_2}.  
These void samples are sometimes combined with auxiliary datasets to train OOD detectors jointly.  
However, this strategy can be problematic because void regions are typically dominated by a limited set of repetitive structures, such as rubbish bins, parking meters and lamps.
Using them as auxiliary OOD data may cause overfitting and reduce the generalization ability of the detector.  

An effective auxiliary OOD sample should be diverse and should not systematically correspond to any specific semantic class.  
In our framework, Perlin Noise-based geometric perturbation is employed to generate diverse and unbiased auxiliary OOD samples.  
This encourages the detector to learn class-agnostic decision boundaries instead of memorizing a few specific object types.  

Therefore, our OOD training objective comprises three components.
First, ID points are optimized using the standard cross-entropy loss for classification.
Second, both Perlin-generated auxiliary OOD samples and ID samples are trained under a binary classification objective.
The objective is formulated as
\begin{align}
\mathcal{L}_{\text{STD}}
&= \mathbb{E}_{x \sim \mathcal{D}_{\text{in}}}
\Big[-\log\!\Big(\frac{e^{S_{\text{NDP}}(f_{\Theta},x)+b}}{1 + e^{S_{\text{NDP}}(f_{\Theta},x)+b}}\Big)\Big] \nonumber\\
&\quad + \,\mathbb{E}_{x \sim \mathcal{D}_{\text{aux}}}
\Big[-\log\!\Big(\frac{1}{1 + e^{S_{\text{NDP}}(f_{\Theta},x)+b}}\Big)\Big],
\label{eq:ee-logistic}
\end{align}
where $b$ is a trainable bias term, $\mathcal{D}_{\text{in}}$ and $\mathcal{D}_{\text{aux}}$ denote the in-distribution 
and Perlin-generated auxiliary OOD data.

The last component introduces a Soft Outlier Exposure (SOE) strategy that handles the void regions present in the training data, allowing the model to utilize them without overfitting to specific void classes.

For ID points, the target probability is close to zero, while for void points, 
the target is a soft label $\beta \in [0,1]$.  
The training objective is defined as:
\begin{align}
\mathcal{L}_{\text{SOE}}
&= \mathbb{E}_{x \sim \mathcal{D}_{\text{in}}}
\big|\sigma(S_{\text{NDP}}(x)+b) \big| \nonumber \\
& + \mathbb{E}_{x \sim \mathcal{D}_{\text{void}}}
\max(0,\beta - \sigma(S_{\text{NDP}}(x)+b)),
\label{eq:sod}
\end{align}
where $\sigma(\cdot)$ denotes the sigmoid function to map the OOD score to probability and $b$ is a trainable bias term. Unless otherwise specified, $\beta$ is set to a fixed value (0.9) in all experiments.

This soft regression design encourages the detector to output low scores for ID regions 
and intermediate scores for uncertain areas, 
preventing overconfidence and improving calibration in complex LiDAR scenes.

The overall training objective for OOD detection involves three terms.
The closed-set term $\mathcal{L}_{\text{CE}}$ uses cross-entropy for ID supervision,  
$\mathcal{L}_{\text{STD}}$ trains with Perlin-generated pseudo-OOD samples,  
and $\mathcal{L}_{\text{SOE}}$ applies soft regularization on void regions.  
The total OOD loss is defined as $\mathcal{L}_{\text{OOD}}
= \mathcal{L}_{\text{CE}}
+ \mathcal{L}_{\text{STD}}
+ \mathcal{L}_{\text{SOE}}.$
For the transformer decoder branch, it follows the closed-set training protocol~\cite{yilmaz2024mask4former,nekrasov2025stu} to maintain ID segmentation performance.

\section{Experiments}

\subsection{Datasets}
\paragraph{Spotting the Unexpected (STU)}
The STU benchmark~\cite{nekrasov2025stu} is a large-scale dataset designed for anomaly segmentation in 3D LiDAR data.
It consists of 72 driving sequences captured with a 128-beam LiDAR sensor and includes both naturally occurring anomalies in real traffic scenes and deliberately placed OOD objects such as buckets, chairs, and surfboards.
Each point is labeled as either \emph{inlier}, \emph{anomaly}, or \emph{unlabeled}.
The dataset provides 19 sequences for validation and 51 for testing, along with 2 sequences reserved for closed-set training and validation.

\begin{figure*}[ht]
    \centering
    \includegraphics[width=\linewidth]{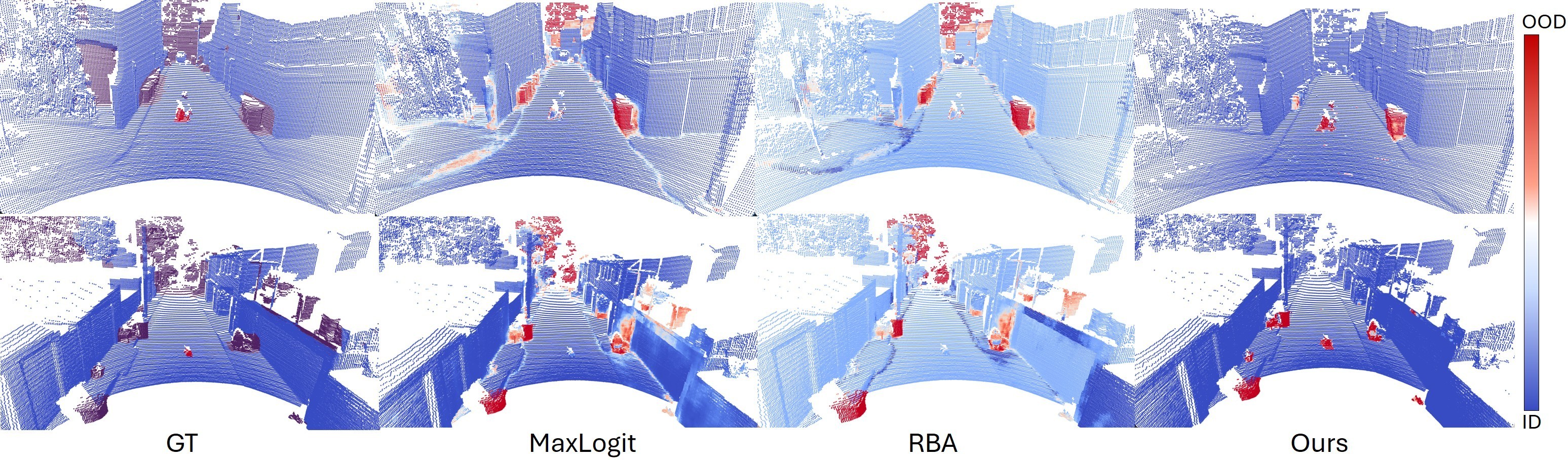}
    \caption{Visualization of OOD score map on the STU benchmark. 
Points are labeled as \textcolor[rgb]{0,0,0.9}{inlier}, \textcolor[rgb]{0.8,0,0}{anomaly}, and \textcolor[rgb]{0.4,0,0.4}{unlabeled}.  
A continuous color bar indicates the predicted likelihood of each point being \textcolor[rgb]{0,0,0.9}{ID} or \textcolor[rgb]{0.8,0,0}{OOD}.  
Our approach yields precise and coherent anomaly masks while maintaining a low false-positive rate on inlier regions. }
    \label{fig:vis_main}
    \vspace{-0.3cm}
\end{figure*}

\paragraph{SemanticKITTI}
We also assess our approach on the SemanticKITTI benchmark~\cite{behley2019semantickitti}, a standard dataset for LiDAR semantic segmentation.
To emulate the OOD detection scenario, we treat categories like \emph{other-structure} and \emph{other-object} as OOD classes, while the remaining categories are considered in-distribution.

\subsection{Evaluation Protocol}
Following the protocol of Nekrasov~\etal~\cite{nekrasov2025stu}, we report both point-level and object-level metrics.
Point-level performance is measured by AUROC, FPR@95, and AP.
For object-level evaluation, panoptic segmentation metrics are used, reporting Recall Quality (RecallQ), Segmentation Quality (SQ), Recognition Quality (RQ), Panoptic Quality (PQ), and Unknown Quality (UQ)~\cite{nekrasov2025stu,wong2019osis,kirillov2019panoptic}.

\subsection{OOD Detection Results}

The STU benchmark focuses on anomaly segmentation for road hazards and provides both point-level and object-level annotations for quantitative evaluation.  
This task can be formulated as point-wise OOD detection, where each LiDAR point is assigned an OOD confidence score.  
We then apply DBSCAN clustering~\cite{dbscan} to the detected OOD points to obtain object-level predictions.

As summarized in \cref{tab:val-results,tab:test-results}, our method achieves state-of-the-art performance in both the validation and \emph{hidden test sets} of the STU benchmark, as verified by the benchmark authors. 
Among all the reported metrics, \emph{Average Precision (AP)} and \emph{Panoptic Quality (PQ)} are the most informative for assessing anomaly segmentation. 
The proposed NDP framework yields consistent improvements across all OOD scoring functions. 
In particular, NDP-EE attains an AP of 74.24\%, representing a 995\% improvement over the previous best result on the STU validation set.
On the hidden test set, NDP-EE maintains strong generalization, achieving 61.31\% AP and 24.99\% PQ, corresponding to more than a $10\times$ improvement in point-level performance and nearly a $2\times$ improvement in object-level performance compared to the previous best method.

\begin{table}[t]
    \centering
    \footnotesize
    \caption{Outlier detection performance on semanticKITTI outlier classes. Object-level performance is not applicable because outlier classes do not have instance masks. * denotes model using Cylinder3D~\cite{zhou2020cylinder3d} architecture. Other methods use the Mask4Former~\cite{yilmaz2024mask4former} architecture.}
    \vspace{-0.1cm}
    \begin{tabular}{lcrrr}
        \toprule
        Method & Aux. Data & AUROC~$\uparrow$ & FPR@95~$\downarrow$ & AP~$\uparrow$ \\
        \midrule
        MaxLogit~\cite{hendrycks2022streethazards} & \xmark & 90.73 & 47.63 & 53.86\\
        RbA~\cite{nayal2023rba} & \xmark & 78.86 & 100.00 & 55.23 \\
        UEM~\cite{nayal2024likelihood} & \cmark & 93.15 & 37.07 & 61.73 \\
        REAL*~\cite{cen2021real} & \cmark & 84.90 & - & 20.08 \\
        APF*~\cite{apf2023} & \xmark & 85.60 & - & 36.10 \\
        LiON*~\cite{lion2025} & \cmark & 92.69 & - & 44.68 \\
        \midrule
        NDP-Energy & \cmark & \textbf{98.32} & \textbf{8.41} & 69.54  \\
        NDP-Entropy & \cmark & 97.93 & 10.15 & 64.06 \\
        NDP-EE & \cmark & 98.01 & 10.38 & \textbf{70.12} \\
        \bottomrule
    \end{tabular}
    \label{tab:semantic_KITTI_ood}
    \vspace{-0.3cm}
\end{table}

We further evaluate our framework on the SemanticKITTI for OOD detection.
As shown in \cref{tab:semantic_KITTI_ood}, the NDP framework consistently boosts the performance of multiple static OOD scoring functions, achieving higher AUROC and lower FPR@95 than all competing methods.  
In particular, NDP-EE attains 70.12\% AP, demonstrating strong cross-dataset generalization.  
Although our method substantially reduces FPR@95, the improvement in AP is modest because SemanticKITTI is not designed for OOD detection. Its outlier classes have limited diversity and do not provide a sufficiently challenging evaluation setting for previous methods.


These results indicate that the NDP framework effectively improves OOD detection performance under severe class imbalance in LiDAR perception.

\begin{table}[htbp]
\centering
\caption{In-distribution panoptic segmentation on the validation sets of STU~\cite{nekrasov2025stu} and SemanticKITTI~\cite{behley2019semantickitti}. Our method outperforms other approaches that involve OOD training.}
\vspace{-0.1cm}
\label{tab:mean-pq}
\footnotesize
\begin{tabular}{lcc}
\toprule
Method & STU (PQ) & SemanticKITTI (PQ) \\
\midrule
Mask4Former-closed-set~\cite{marcuzzi2023maskpls} & 52.73 & 60.72 \\
Mask4Former-void~\cite{blum2021fishyscapes} & 26.96 & 47.97 \\
Mask4Former-NDP & 52.37 & 59.38 \\
\bottomrule
\end{tabular}
\vspace{-0.3cm}
\end{table}

\subsection{In-distribution Segmentation Results}
An effective OOD detection framework should preserve the model’s capability to segment in-distribution classes. 
We therefore evaluate the closed-set panoptic segmentation performance of our approach on the validation sets of both STU and SemanticKITTI. 
As shown in \cref{tab:mean-pq}, our model (NDP-EE) retains comparable segmentation accuracy relative to the standard Mask4Former baseline. 
On STU, Mask4Former-NDP attains a PQ of 52.37, matching the closed-set performance of Mask4Former while substantially outperforming variants trained with void classification.
Similarly, on SemanticKITTI, Mask4Former-NDP maintains high segmentation quality with a PQ of 59.38, closely aligning with the original closed-set Mask4Former and exceeding other OOD-training-based counterparts. 
These results confirm that the integration of the proposed NDP module does not degrade the closed-set performance of the model.

\subsection{Quantitative Results}
\Cref{fig:vis_main} visualizes the OOD uncertainty maps predicted by our NDP module compared to several baselines from the STU benchmark. For fair visualization, all OOD scores are linearly normalized and clamped to [0,1]. In the ground truth map, inlier regions are shown in blue, anomaly regions in red, and unlabeled regions from STU in dark purple. Based on the observation, baseline methods such as MaxLogit~\cite{hendrycks2022streethazards} and RBA~\cite{nayal2023rba} can only detect some unlabeled regions while failing to identify the anomaly objects. In contrast, our method not only accurately detects the OOD objects but also identifies regions that are outside the inlier distribution of the labeled training data. Compared with the baselines, our method avoids false positive predictions on distant background regions, which further demonstrates the effectiveness of the NDP module.

\begin{table}[t]
    \footnotesize
    \centering
    \caption{Ablation study of the NDP framework. All components contribute positively to the overall performance.}
    \begin{tabular}{lrrr}
        \toprule
        Method & AUROC~$\uparrow$ & FPR@95~$\downarrow$ & AP~$\uparrow$ \\
        \midrule
        Void Training & 94.54 & 22.51 & 2.09 \\
        Mixed Training & 98.47 & 8.28 & 11.03 \\
        w/o SOE & 99.32 & 1.92 & 67.10 \\
        w/o NDP & 99.18 & 3.02 & 58.69 \\
        NDP-EE & \textbf{99.53} & \textbf{1.43} & \textbf{74.24} \\
        \bottomrule
    \end{tabular}
    \label{tab:ablation}
    \vspace{-0.25cm}
\end{table}

\subsection{Ablation Study}
To assess the contribution of each component in our framework, we perform an ablation study of NDP-EE on the validation set of the STU benchmark, as summarized in \cref{tab:ablation}. 
The baseline \emph{Void Training} uses only void classes (points excluded from closed-set categories) as reliable OOD samples, resulting in weak OOD separation. 
Mixed training refers to using both void samples and Perlin-noise-generated OOD samples in $\mathcal{L}_{\text{STD}}$ (Eq. 7), but its performance remains limited.
When the Soft Outlier Exposure loss (\emph{w/o SOE}) is removed and the model is trained solely with Perlin noise–based OOD samples, AP decreases by 7.14\% compared with the full model.
Further eliminating NDP (\emph{w/o NDP}) and relying on a static OOD score leads to a 15.55\% reduction in AP, indicating that NDP is critical for stable optimization and well-calibrated confidence estimation. 
The complete model achieves the best overall results, confirming that NDP, together with the proposed OOD training strategy, provides robust OOD detection under severe class imbalance.

\paragraph{Comparison with Statistical Class Prior}
We further compare the proposed NDP with the statistical class prior reweighting method~\cite{jiang2023classprior} and the regularization method~\cite{choi2023balanced}. For fairness, we adopt the standard energy-based OOD score~\cite{energy} and use the same OE strategy. As shown in \cref{tab:ablation_prior}, methods based on a fixed statistical prior struggle under the severe LiDAR imbalance, where class counts differ by several orders of magnitude. In contrast, the learnable NDP adapts to class-dependent behavior, achieving 99.37\% AUROC, 2.35\% FPR@95, and 66.54\% AP. Using the static OOD function without NDP (w/o NDP) causes a 9.66\% drop in AP. These results indicate the effectiveness of the learnable distribution prior.

\begin{table}[t]
    \centering
    \footnotesize
    \caption{Comparison with the statistic class distribution prior using energy-based OOD score. NDP largely outperforms the statistics class distribution prior and static OOD score.}
    \begin{tabular}{lrrr}
        \toprule
        Method  & AUROC~$\uparrow$ & FPR@95~$\downarrow$ & AP~$\uparrow$ \\
        \midrule
        w/o NDP & 98.50 & 4.33 & 56.88 \\
        Statistics Prior~\cite{jiang2023classprior} & 97.99 & 3.72 & 23.87 \\
        Balanced Energy~\cite{choi2023balanced} & 98.47 & 3.17 & 57.05\\
        NDP-Energy  & \textbf{99.37} & \textbf{2.35} & \textbf{66.54} \\ 
        \bottomrule
    \end{tabular}
    \label{tab:ablation_prior}
    \vspace{-0.2cm}
\end{table}

\begin{table}[t]
    \centering
    \footnotesize
    \caption{Ablation study of $\beta$ in soft outlier exposure. SOE is not sensitive to the choice of hyperparameter.}
    \begin{tabular}{lrrr}
        \toprule
        $\beta$ in SOE  & AUROC~$\uparrow$ & FPR@95~$\downarrow$ & AP~$\uparrow$ \\
        \midrule
        0.7 & 99.26 & 1.53 & 68.00 \\
        0.8 & 99.36 & 2.35 & 72.73 \\
        0.9 & 99.53 & 1.43 & 74.24 \\
        1.0 & 99.36 & 1.61 & 73.06 \\ 
        \bottomrule
    \end{tabular}
    \label{tab:ablation_soe}
    \vspace{-0.3cm}
\end{table}

\paragraph{Effect of $\beta$ in Soft Outlier Exposure}
We further study the influence of the hyperparameter $\beta$ in the SOE loss, which controls the confidence level assigned to soft OOD labels.  
A smaller $\beta$ makes OOD supervision weaker and less distinguishable from ID samples, while a larger value approximates hard labeling and may overfit to void regions.  
As shown in \cref{tab:ablation_soe}, performance remains across a broad range of $\beta$ values, confirming the robustness of SOE.

\subsection{Limitation}
While our approach substantially improves LiDAR OOD detection, challenges remain due to the inherently irregular structure of point clouds. OOD objects exhibit diverse and unpredictable geometries. Unlike closed-set segmentation, there are no ground-truth OOD samples available for supervised boundary learning, which constrains boundary accuracy. As shown in \cref{fig:vis_main}, this can lead to small discrepancies between predicted OOD boundaries and annotated anomalies.
Future work will explore advanced post-processing strategies to refine point-level OOD masks and enhance boundary precision.

\section{Conclusion}
We presented the Neural Distribution Prior (NDP), a learnable framework for robust LiDAR OOD detection.
By modeling class-dependent predictive distributions and reweighting OOD scores through a lightweight attention mechanism, NDP effectively alleviates the impact of severe class imbalance in LiDAR data.
Combined with the Soft Outlier Exposure (SOE) strategy and Perlin noise–based OOD synthesis, our method achieves state-of-the-art performance on the STU and SemanticKITTI benchmarks while maintaining closed-set segmentation accuracy.
Future work will extend OOD detection to multimodal and spatiotemporal features for more generalizable open-world perception. We hope this study will inspire further research in this direction.

\section*{Acknowledgments}
\noindent The first two authors acknowledge the financial support from The University of Melbourne through the Melbourne Research Scholarship. Feng Liu is supported by the Australian Research Council (ARC) with the grant
number DE240101089. This research was supported by The University of Melbourne’s Research Computing Services and the Petascale Campus Initiative.
{
    \small
    \bibliographystyle{ieeenat_fullname}
    \bibliography{main}
}
\clearpage
\setcounter{page}{1}
\maketitlesupplementary

\section{Implementation Details}

We initialize the model using a Mask4Former checkpoint pretrained on SemanticKITTI~\cite{behley2019semantickitti} and Panoptic CUDAL~\cite{tseng2025panopticcudal}. The model is then fine-tuned for up to 10 epochs on the downstream datasets, with Perlin noise–synthesized OOD samples included during training. Optimization uses AdamW with a learning rate of $2\times10^{-4}$ and a batch size of 8 on NVIDIA A100 GPUs.
For the NDP matrix $\psi$, the embedding dimension $d$ is set to 16 unless indicated otherwise. In the SOE loss, the soft OOD target $\beta$ is fixed to 0.9 for all experiments unless otherwise specified.
To compensate for the scarcity of auxiliary OOD points compared to in-distribution points, their loss contribution is weighted 10000 times higher than that of ID points.

For the Perlin Raise algorithm, the patch radius $r$ is sampled from $[0.75, 1.5]$, the noise strength $\alpha$ is set to 0.4, and the target ratio $\rho$ is fixed at 0.3.

\section{Explanation of Evaluation Metrics}
\paragraph{Point-level Evaluation Metrics}
Point-level evaluation metrics for LiDAR OOD detection include AUROC, FPR@95, and Average Precision (AP). These metrics are widely used in OOD detection and anomaly segmentation~\cite{yang2024generalizedood,chan2021segmentmeifyoucan,fishyscapes_2,nekrasov2025stu}.

\textbf{AUROC} assesses how well the OOD score separates OOD points from ID points across all possible thresholds. It is obtained by ranking points by their OOD scores and measuring how consistently OOD points receive higher scores than ID points. Because it is threshold-free, AUROC reflects the overall separability of the score function. However, this metric is not ideal for scenarios with severe ID/OOD imbalance, and AP is therefore often used as the main evaluation metric~\cite{fishyscapes_2,nekrasov2025stu}.

\textbf{FPR@95} measures the reliability of the detector at a high-recall operating point. We first determine the score threshold that correctly identifies 95\% of OOD points, and then evaluate the proportion of ID points incorrectly flagged as OOD at this threshold.

\textbf{Average Precision} evaluates the quality of OOD detection under the precision-recall trade-off. By sweeping the score threshold from high to low, AP quantifies how well the detector maintains high precision as it covers more OOD points. The AP score is the integral of the resulting precision–recall curve, typically approximated through monotonic interpolation. AP is especially informative for LiDAR OOD segmentation because it naturally handles the severe imbalance between ID and OOD points. 

\paragraph{Object-level Evaluation Metrics}
The STU benchmark~\cite{nekrasov2025stu} provides fine-grained instance masks for all OOD objects and adopts Panoptic Quality (PQ)~\cite{kirillov2019panoptic} as the primary metric for object-level anomaly segmentation. PQ evaluates instance-level performance by combining Segmentation Quality (SQ) and Recognition Quality (RQ). For a class $c$, it is defined as:
\resizebox{0.95\linewidth}{!}{
\begin{minipage}{\linewidth}
\begin{equation}
\text{PQ}_c
=
\underbrace{\frac{\sum_{(p, g) \in TP_c} \mathrm{IoU}(p, g)}{|TP_c|}}_{\text{Segmentation Quality (SQ)}}
\times
\underbrace{\frac{|TP_c|}{|TP_c| + \frac{1}{2}|FP_c| + \frac{1}{2}|FN_c|}}_{\text{Recognition Quality (RQ)}}.
\end{equation}
\end{minipage}
}

A predicted object is counted as a true positive (TP) if it overlaps with a ground-truth instance with Intersection over Union (IoU) greater than $50\%$. Unmatched predictions are counted as false positives (FP), and missed ground-truth instances as false negatives (FN). Ignore regions are excluded from evaluation and predictions inside these regions are not penalized. 

For in-distribution classes, the final PQ score is obtained by averaging $\mathrm{PQ}_c$ over all classes. For anomaly segmentation, all OOD objects are grouped into a single class, and PQ is reported for this aggregated category.

To quantify anomaly recall, STU also reports the Unknown Quality (UQ) metric~\cite{wong2019osis}:
\begin{equation}
\text{UQ}
=
\underbrace{\frac{\sum_{(p, g) \in TP} \mathrm{IoU}(p, g)}{|TP|}}_{\text{Segmentation Quality (SQ)}}
\times
\underbrace{\frac{|TP|}{|TP| + |FN|}}_{\text{Recall Quality (RecallQ)}}.
\end{equation}
Unlike PQ, UQ does not penalize false positives, allowing the metric to focus purely on the model's ability to retrieve anomaly instances. As with PQ, an IoU threshold of $50\%$ is required to count a prediction as a true positive. However, anomaly segmentation in LiDAR scenes requires both high anomaly recall and careful control of false positives, since excessive false alarms can negatively affect downstream planning~\cite{nekrasov2025stu}.


\begin{figure*}[ht]
    \centering
    \includegraphics[width=\linewidth]{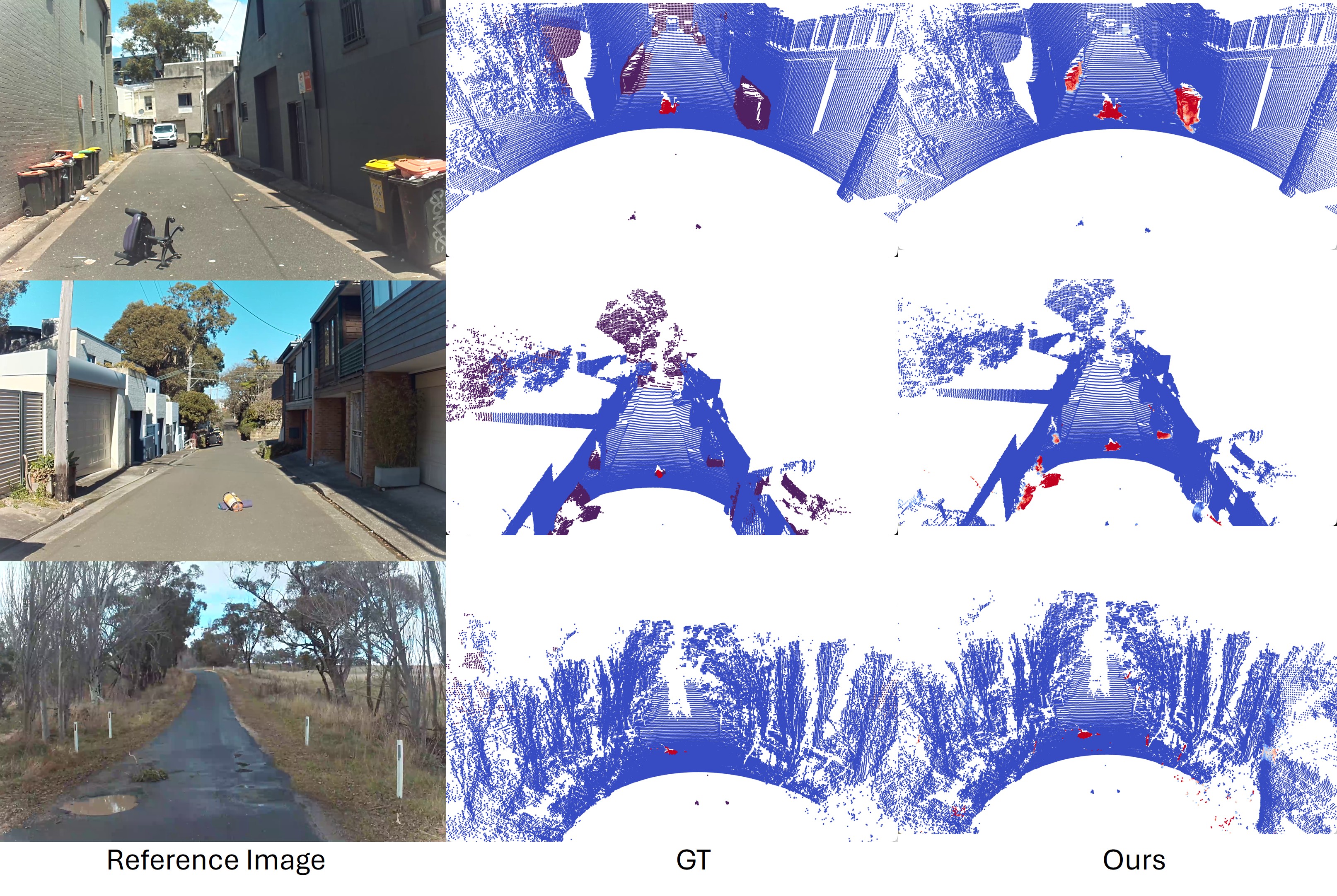}
    \caption{Additional visualization of OOD score map on the STU benchmark with image reference. Points are labeled as \textcolor{blue}{inlier}, \textcolor{red}{anomaly}, and \textcolor[rgb]{0.4,0,0.4}{unlabeled}. Our approach yields precise and coherent anomaly masks while maintaining a low false-positive rate on inlier regions.}
    \label{fig:vis_img}
\end{figure*}

\section{Additional Visualization}
\begin{figure*}[ht]
    \centering
    \includegraphics[width=\linewidth]{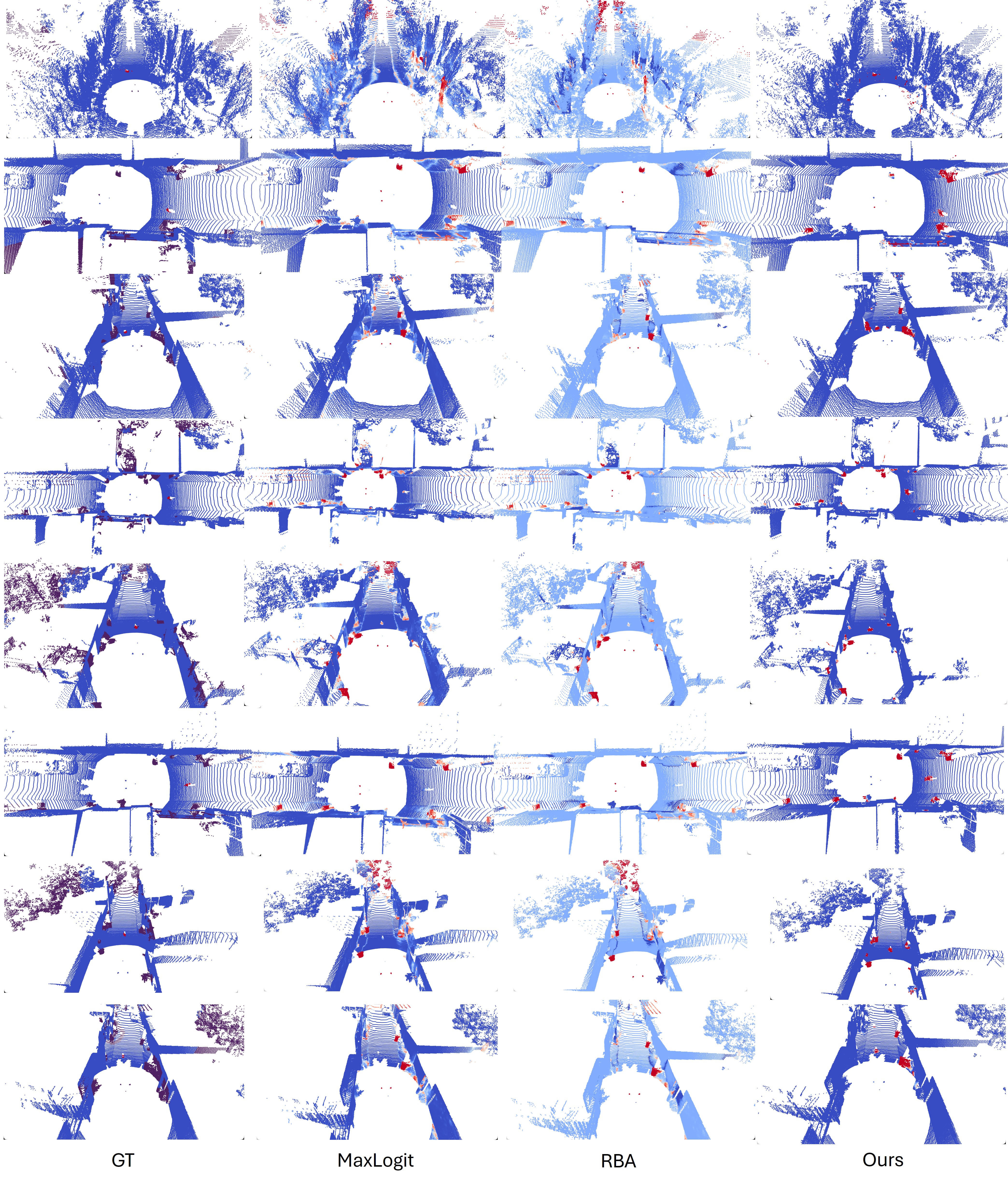}
    \caption{Additional visualization of OOD score map on the STU benchmark.  
Points are labeled as \textcolor{blue}{inlier}, \textcolor{red}{anomaly}, and \textcolor[rgb]{0.4,0,0.4}{unlabeled}.
Our approach yields precise and coherent anomaly masks while maintaining a low false-positive rate on inlier regions. }
    \label{fig:vis_suppl}
\end{figure*}

\cref{fig:vis_img} and \cref{fig:vis_suppl} illustrate the qualitative performance of our method. Across diverse environments, including narrow urban alleys and unstructured rural roads, the model consistently identifies a broad range of OOD objects such as armchairs, fallen branches, packages, and yoga mats. Our method also substantially reduces the false positive rate. In addition, baseline approaches such as MaxLogit~\cite{hendrycks2022streethazards} and RbA~\cite{nayal2023rba} incorrectly label tree trunks as OOD in forest environments, where dense geometry and cluttered backgrounds make boundary estimation difficult. Our model maintains reliable predictions in these complex scenes, avoiding such false positives and producing cleaner and more consistent OOD masks under challenging structural variability.

\begin{figure*}[ht]
    \centering
    \includegraphics[width=\linewidth]{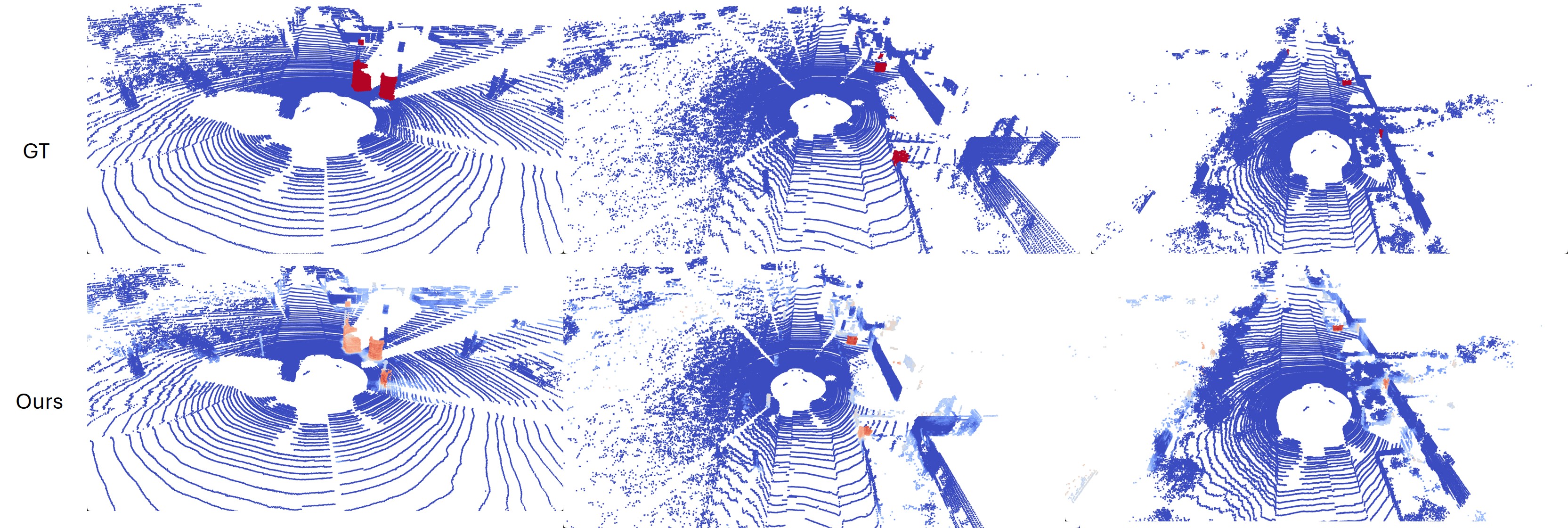}
    \caption{Visualization of OOD score map on the SemanticKITTI.  
Points are labeled as \textcolor{blue}{inlier}, \textcolor{red}{anomaly}, and \textcolor[rgb]{0.4,0,0.4}{unlabeled}.
Our approach yields precise and coherent anomaly masks while maintaining a low false-positive rate on inlier regions. }
    \label{fig:sk_vis}
\end{figure*}

In addition, as showin in \cref{fig:sk_vis}, we provide visualizations of the OOD map on SemanticKITTI, where our method still performs well, demonstrating generalization across datasets.

\section{Additional Results}

\cref{tab:ablation_d} presents an ablation study on the template size $d$ of the NDP matrix $\psi$, where $d$ determines the dimensionality of the vectors stored in $\psi$ as the learnable prior.
A moderate NDP size yields the best performance: $d=16$ achieves the highest AP (74.24\%) and a strong AUROC (99.53\%).  
Overall, NDP is not highly sensitive to this hyperparameter.  
Smaller values of $d$ store fewer parameters and struggle to capture the dynamics of the logit distribution, whereas larger values introduce additional parameters that are more difficult to optimize and may lead to overfitting.

\begin{table}[ht]
    \centering
    \footnotesize
    \caption{Ablation study of template size $d$ in NDP matrix $\psi$, where $d$ is the dimensionality of the vectors stored in $\psi$ as the learnable prior.}
    \begin{tabular}{lrrr}
        \toprule
        $d$ & AUROC~$\uparrow$ & FPR@95~$\downarrow$ & AP~$\uparrow$ \\
        \midrule
        8 & 99.42 & 1.21 & 70.29 \\
        16 & \textbf{99.53} & \textbf{1.43} & \textbf{74.24} \\
        32 & 99.20 & 1.67 & 70.14 \\
        \bottomrule
    \end{tabular}
    \label{tab:ablation_d}
\end{table}

We validated our method using a lightweight MinkUNet~\cite{choy2019minkowski} backbone. As shown in \cref{tab:mink}, our method consistently improves OOD detection performance.

\begin{table}[ht]
    \centering
    \caption{OOD Detection Results of NDP-EE using various backbones}
    \footnotesize
    \begin{tabular}{lrrr}
        \toprule
           Method & AUROC~$\uparrow$ & FPR@95~$\downarrow$ & AP~$\uparrow$ \\
        \midrule
        Static Extended Energy & 98.33 & 2.94 & 58.36 \\
        NDP-EE & \textbf{99.19} & \textbf{2.89} & \textbf{70.29} \\ 
        \bottomrule
    \end{tabular}
    \label{tab:mink}
\end{table}

\begin{table*}[htbp]
\setlength\tabcolsep{3.7pt}
    \begin{center}
    \caption{In-distribution per-class performance of the methods on the validation set of STU~\cite{nekrasov2025stu} dataset. Our model retains comparable panoptic segmentation performance to standard Mask4Former training.}
        \resizebox{\textwidth}{!}{
        \label{tab:supp-results-inlier}
        \footnotesize
        \begin{tabular}{l|c|cccccccccccccc|c}
            \toprule
            Method &
            \begin{sideways}void\end{sideways} &
            \begin{sideways}car\end{sideways} &
            \begin{sideways}truck\end{sideways} &
            \begin{sideways}bicycle\end{sideways} &
            \begin{sideways}person\end{sideways} &
            \begin{sideways}road\end{sideways} &
            \begin{sideways}sidewalk\end{sideways} &
            \begin{sideways}parking\end{sideways} &
            \begin{sideways}building\end{sideways} &
            \begin{sideways}vegetation\end{sideways} &
            \begin{sideways}trunk\end{sideways} &
            \begin{sideways}terrain\end{sideways} &
            \begin{sideways}fence\end{sideways} &
            \begin{sideways}pole\end{sideways} &
            \begin{sideways}traffic sign\end{sideways} &
            PQ \\
            \midrule
            Mask4Former~\cite{yilmaz2024mask4former} & -- & 80.99 & 37.28 & 47.65 & 80.99 & 71.46 & 17.74 & 0.0 & 84.08 & 89.73 & 29.34 & 30.79 & 47.6 & 59.62 & 60.96 & 52.73 \\
            Mask4Former-void~\cite{blum2021fishyscapes} & 0.07 & 23.88 & 20.78 & 1.01 & 43.30 & 38.24 & 20.03 & 11.11 & 48.45 & 43.09 & 20.20 & 17.31 & 30.80 & 27.26 & 33.16 & 26.96 \\
            \hline
            Mask4Former-NDP & -- & 77.42 & 48.58 & 51.47 & 76.05 & 40.37 & 12.37 & 0.0 & 90.83 & 92.84 & 31.00 & 65.09 & 36.14 & 51.28 & 59.77 & 52.37\\
            \bottomrule
        \end{tabular}
        }
    \end{center}
    
\end{table*}

\begin{table*}[!t]
\setlength\tabcolsep{3.7pt}
    \begin{center}
    \caption{In-distribution per-class performance of the methods on validation sets of SemanticKITTI~\cite{behley2019semantickitti}. Our model retains comparable panoptic segmentation performance to standard Mask4Former training.}
        \resizebox{\textwidth}{!}{
        \label{tab:supp-results-semKITTI}
        \footnotesize
        \begin{tabular}{l|c|ccccccccccccccccccc|c}
            \toprule
            Method &
            \begin{sideways}void\end{sideways} &
            \begin{sideways}car\end{sideways} &
            \begin{sideways}truck\end{sideways} &
            \begin{sideways}bicycle\end{sideways} &
            \begin{sideways}motorcycle\end{sideways} &
            \begin{sideways}other vehicle\end{sideways} &
            \begin{sideways}person\end{sideways} &
            \begin{sideways}bicyclist\end{sideways} &
            \begin{sideways}motorcyclist\end{sideways} &
            \begin{sideways}road\end{sideways} &
            \begin{sideways}sidewalk\end{sideways} &
            \begin{sideways}parking\end{sideways} &
            \begin{sideways}other ground\end{sideways} &
            \begin{sideways}building\end{sideways} &
            \begin{sideways}vegetation\end{sideways} &
            \begin{sideways}trunk\end{sideways} &
            \begin{sideways}terrain\end{sideways} &
            \begin{sideways}fence\end{sideways} &
            \begin{sideways}pole\end{sideways} &
            \begin{sideways}traffic sign\end{sideways} &
            PQ \\
            \midrule
            Mask4Former~\cite{yilmaz2024mask4former} & -- & 93.53 & 59.39 & 62.55 & 64.82 & 54.36 & 79.61 & 89.16 & 25.01 & 93.24 & 77.90 & 28.79 & 0.0 & 87.27 & 87.28 & 51.08 & 59.92 & 24.85 & 56.76 & 58.14 & 60.72\\
            Mask4Former-void~\cite{blum2021fishyscapes} & 6.08 & 74.36 & 47.00 & 32.19 & 43.34 & 33.30 & 42.90 & 68.75 & 00.33 & 93.35 & 77.07 & 19.01 & 0.0 & 82.77 & 81.34 & 47.56 & 56.94 & 19.98 & 54.48 & 36.82 & 47.97 \\
            \hline
            Mask4Former-NDP & -- & 93.22 & 60.77 & 60.17 & 68.99 & 56.62 & 80.25 & 87.93 & 0.0 & 92.90 & 77.29 & 25.44 & 0.0 & 87.28 & 86.88 & 52.01 & 59.89 & 23.61 & 58.31 & 56.73 & 59.38 \\

            \bottomrule
        \end{tabular}
        }
    \end{center}
    
\end{table*}

\cref{tab:supp-results-inlier} and \cref{tab:supp-results-semKITTI} report the in-distribution per-class panoptic segmentation performance on the STU and SemanticKITTI validation sets. Our model (NDP-EE) preserves segmentation accuracy comparable to the standard Mask4Former~\cite{yilmaz2024mask4former} baseline.
On STU~\cite{nekrasov2025stu}, Mask4Former-NDP achieves a PQ of 52.37, matching the closed-set performance of Mask4Former and substantially surpassing variants trained with void classification.
On SemanticKITTI~\cite{behley2019semantickitti}, Mask4Former-NDP maintains strong segmentation quality with a PQ of 59.38, closely tracking the original closed-set Mask4Former and outperforming other OOD-training-based counterparts.
These results indicate that the incorporation of the proposed NDP module does not compromise closed-set segmentation performance.

\begin{figure*}[htbp]
    \centering
    \begin{subfigure}[t]{\textwidth}
        \centering
        \includegraphics[width=\textwidth]{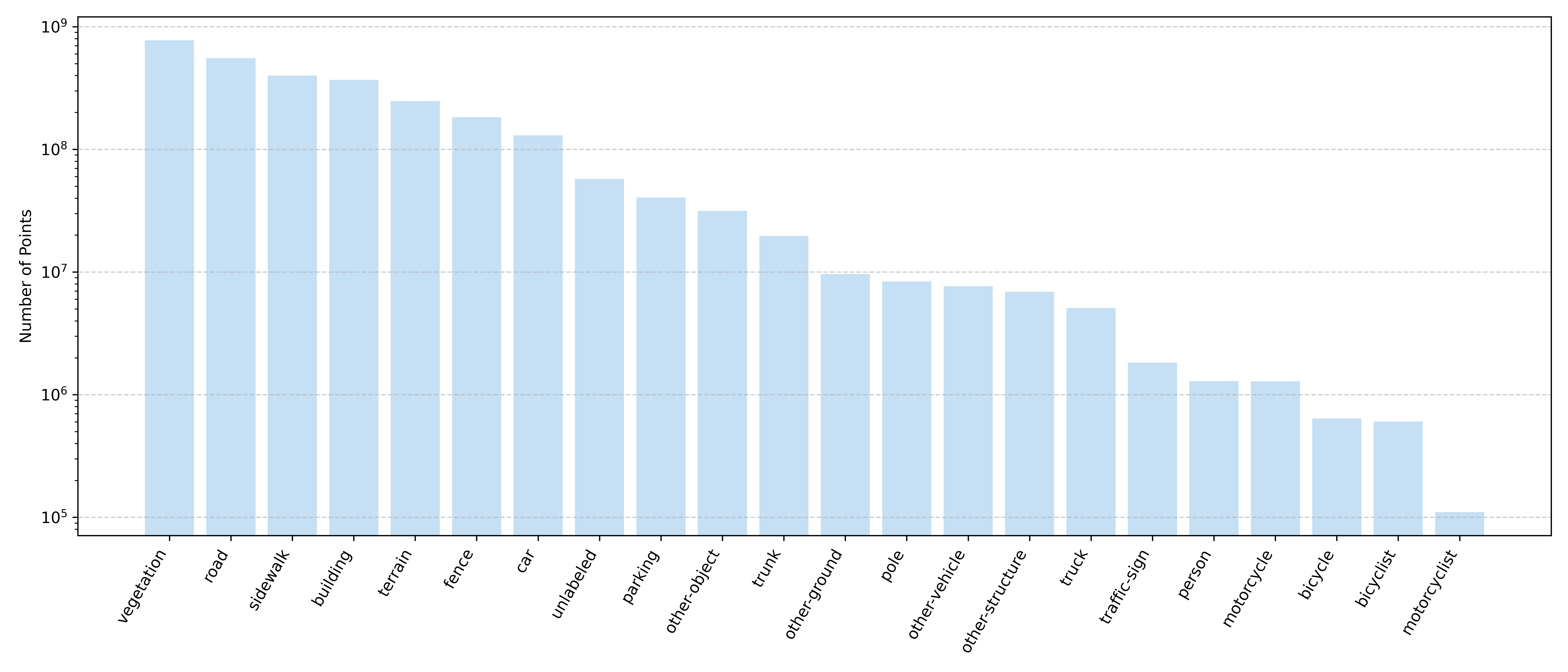}
        \caption{Per-class point counts (log scale).}
        \label{fig:sk-class-counts}
    \end{subfigure}
    \hfill
    \begin{subfigure}[t]{\textwidth}
        \centering
        \includegraphics[width=\textwidth]{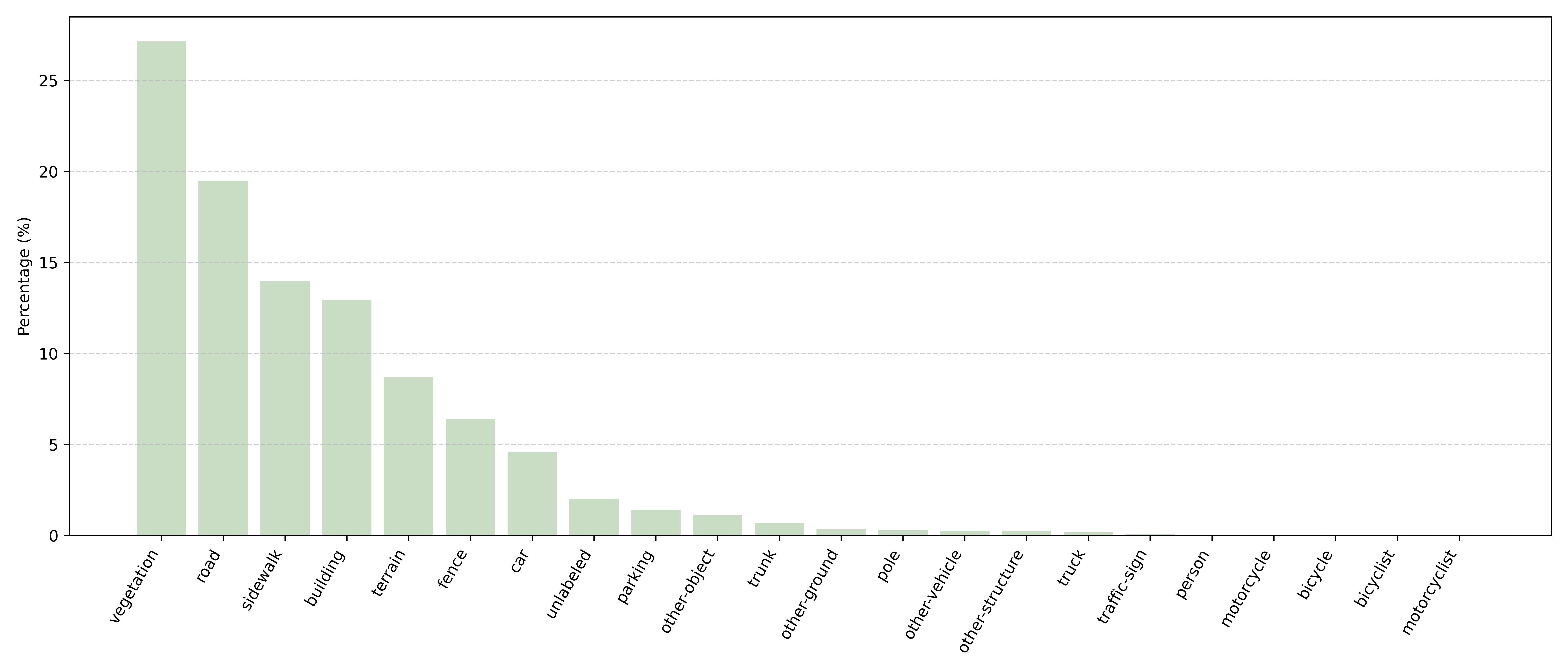}
        \caption{Per-class point percentage.}
        \label{fig:sk-class-percent}
    \end{subfigure}
    \caption{
        Class distribution in the SemanticKITTI dataset. In the dataset, vegetation, road, and sidewalk account for most points. Vegetation alone contributes more than a quarter of all points, and the top four to five classes collectively exceed half of the dataset. In contrast, many classes such as motorcyclist, bicyclist, bicycle, person, and traffic sign appear in extremely small proportions. These categories often fall below one percent of the total point count.
    }
    \label{fig:sk-class-imbalance}
\end{figure*}

\begin{figure*}[htbp]
    \centering
    \begin{subfigure}[t]{\textwidth}
        \centering
        \includegraphics[width=\textwidth]{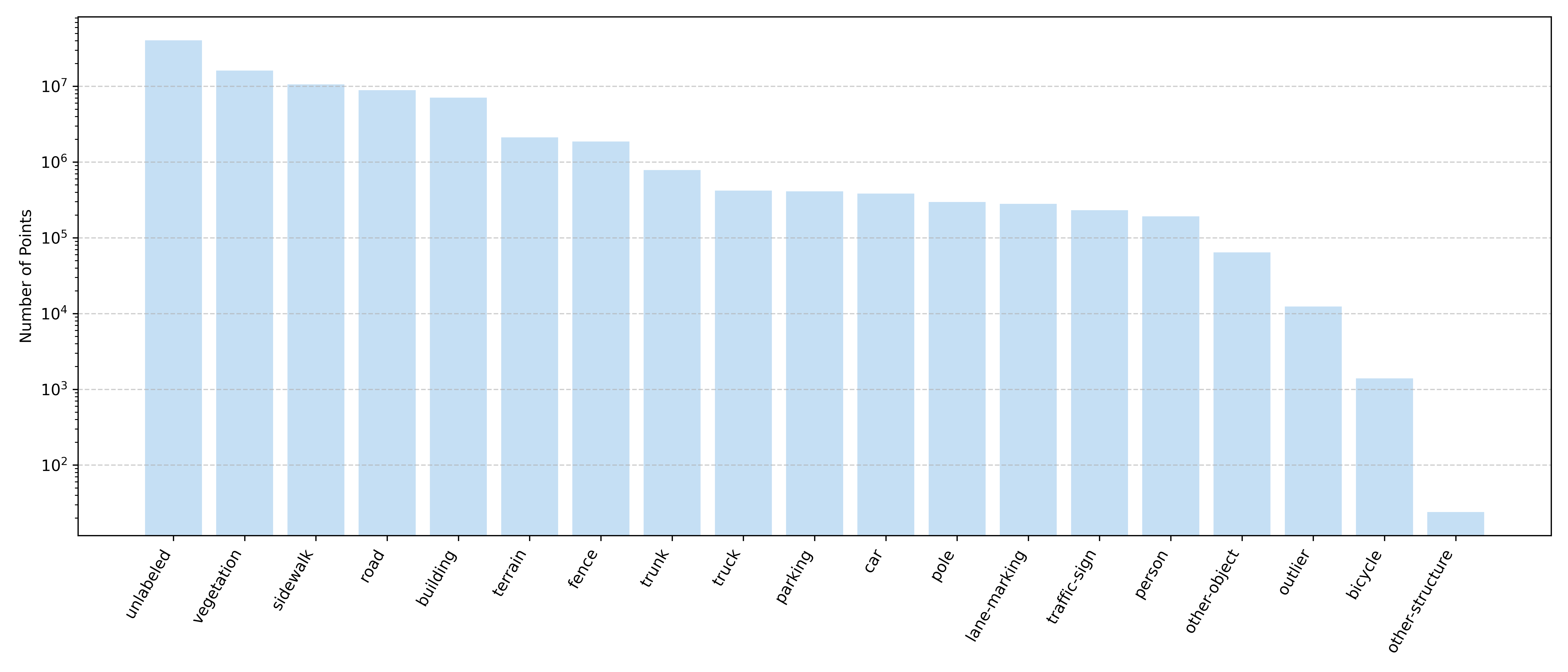}
        \caption{Per-class point counts (log scale).}
        \label{fig:stu-class-counts}
    \end{subfigure}
    \hfill
    \begin{subfigure}[t]{\textwidth}
        \centering
        \includegraphics[width=\textwidth]{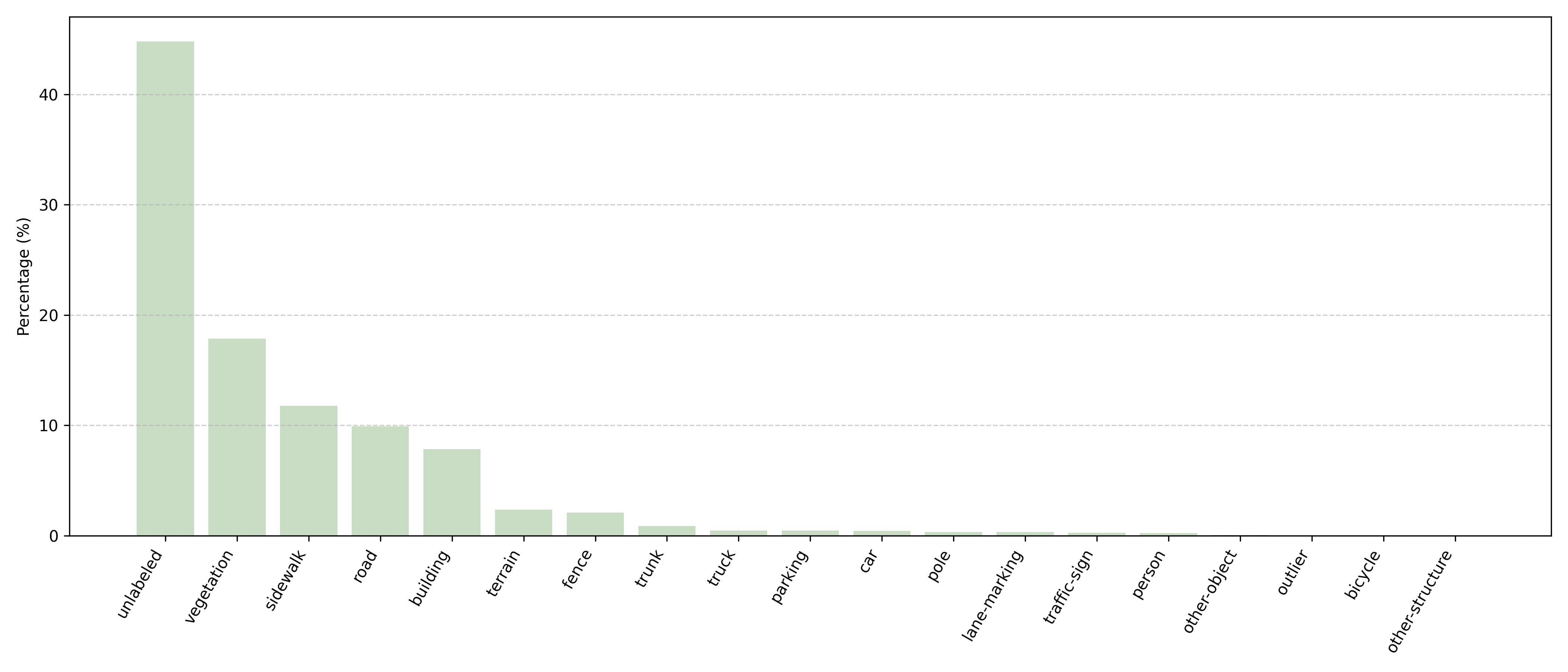}
        \caption{Per-class point percentage.}
        \label{fig:stu-class-percent}
    \end{subfigure}
    \caption{
        Class distribution in the STU dataset. Using the sequence 201 with full annotation as an example, the scene contains over $90$ million LiDAR points in total.
        A few dominant classes, such as vegetation, road, and sidewalk, occupy most of the point cloud.
        Rare classes like person, traffic-sign, and bicycle comprise less than $0.5\%$ of all points.
    }
    \label{fig:stu-class-imbalance}
\end{figure*}

\section{Dataset Statistics}
For OOD detection, class imbalance is especially severe in LiDAR data and makes anomaly discrimination more difficult. This motivates the use of adaptive mechanisms such as distribution-aware priors or dynamic reweighting.

As shown in \cref{fig:sk-class-imbalance}, SemanticKITTI~\cite{behley2019semantickitti} exhibits an extremely long-tailed distribution. Vegetation, road, and sidewalk account for the majority of points. Vegetation alone contributes more than one quarter of the dataset, and the top four to five classes collectively comprise more than half of all annotated points. In contrast, classes such as motorcyclist, bicyclist, bicycle, person, and traffic sign appear in very small quantities, often below one percent of the total point count.

In STU~\cite{nekrasov2025stu}, most evaluation sequences provide only three labels: inlier, anomaly, and unlabeled, without detailed in-distribution class annotations. We therefore use sequence 201, which includes full semantic labels, as a representative example. As shown in \cref{fig:stu-class-imbalance}, this scene contains over 90 million LiDAR points. Similar to SemanticKITTI, a few head classes, including vegetation, road, and sidewalk, dominate the point cloud, while rare categories such as person, traffic sign, and bicycle account for less than 0.5\% of all points.

Our innovation directly targets this issue. By introducing a learnable distribution prior and reweighting logits through a class-dependent attention mechanism, the proposed framework models the characteristic prediction patterns of each class rather than assuming a uniform inlier distribution. This enables more faithful calibration across both head and tail categories and substantially improves OOD scoring in long-tailed LiDAR scenes.

\begin{figure}[t]
    \centering
    \includegraphics[width=\linewidth]{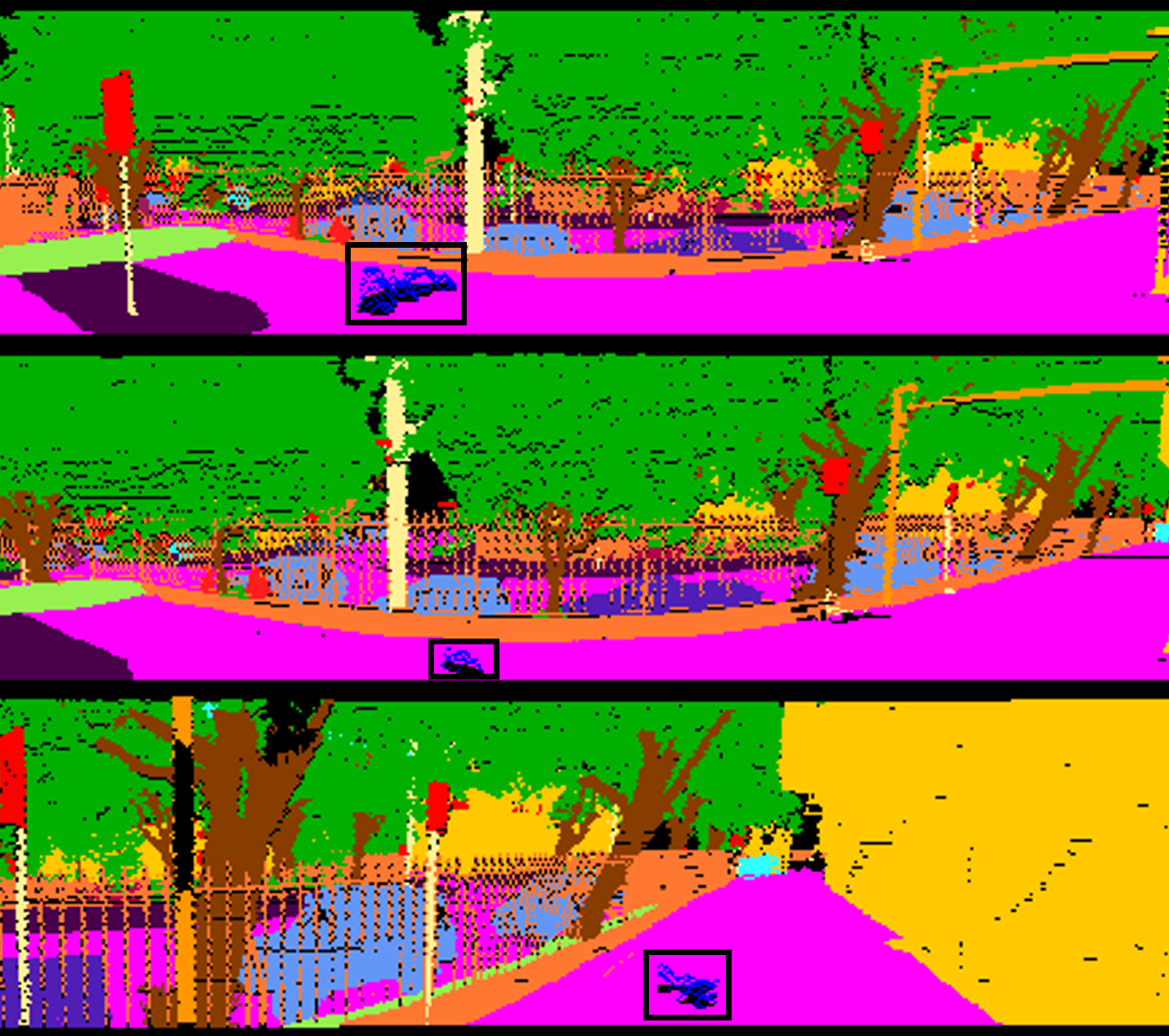}
    \caption{Range-view visualization of Perlin Raise–generated OOD samples. Blue regions denote synthetic anomalies. }
    \label{fig:ood_range}
\end{figure}

\section{Visualization of OOD Samples Generated by Perlin Noise}

The Perlin Raise augmentation produces synthetic OOD regions highlighted in blue, which exhibit substantial variation in geometry and scale. As shown in \cref{fig:ood_range}, these OOD insertions span small localized perturbations to larger, irregular structures that integrate coherently with the surrounding scene layout. This diversity yields a wide range of anomaly shapes that are not repetitive and do not correspond to any in-distribution semantic category. The resulting samples provide a rich and varied training signal for OOD detection, enabling the model to learn more generalizable decision boundaries and reducing susceptibility to overfitting on narrowly defined auxiliary OOD data.

Although Perlin noise does not explicitly model occlusion, we observe that it still performs well in practice. Future work may incorporate more realistic geometric constraints, such as occlusion-aware generation.

\end{document}